%% file: submission.tex
\documentclass[10pt,twocolumn,letterpaper]{article}

\usepackage{./cvpr_stuff/cvpr}
\usepackage{times}
\usepackage{epsfig}
\usepackage{graphicx}
\usepackage{amsmath}
\usepackage{amssymb}
\usepackage{bm}
\usepackage{float}
\usepackage{wrapfig}

\usepackage[breaklinks=true,bookmarks=false]{hyperref}

 \cvprfinalcopy 

\def\cvprPaperID{1244} 

\ifcvprfinal\fi

\usepackage[percent]{overpic}
\hyperbaseurl{https://}

\input{./0_defs.tex}
\begin{document}

\title{Product Manifold Filter: Non-Rigid Shape Correspondence \\via Kernel Density Estimation in the Product Space}

\author{Matthias Vestner\\
Technical University Munich\\
%
\and
Roee Litman\\
Tel-Aviv University\\
%
\and
Emanuele Rodol\`a\\
USI Lugano\\
%
\and
Alex Bronstein\\
Technion, Israel Institute of Technology\\
Perceptual Computing Group, Intel, Israel\\
%
\and
Daniel Cremers\\
Technical University Munich\\
}

\maketitle

\begin{abstract}
Many algorithms for the computation of correspondences between deformable shapes rely on some variant of nearest neighbor matching in a descriptor space. Such are, for example, various point-wise correspondence recovery algorithms used as a post-processing stage in the functional correspondence framework.
Such frequently used techniques implicitly make restrictive assumptions (e.g., near-isometry) on the considered shapes and in practice suffer from lack of accuracy and result in poor surjectivity.
We propose an alternative recovery technique capable of guaranteeing a bijective correspondence and producing significantly higher accuracy and smoothness. Unlike other methods our approach does not depend on the assumption that the analyzed shapes are isometric.
We derive the proposed method from the statistical framework of kernel density estimation and demonstrate its performance on several challenging deformable 3D shape matching datasets.
\end{abstract}

\input{./1_intro.tex}

\input{./2_method.tex}

\input{./4_results.tex}

\input{./5_conc.tex}

{\small
\bibliographystyle{./cvpr_stuff/ieee}
\bibliography{literature}
}

\end{document}

%% file: 0_defs.tex
\newcommand{\Rr}{\mathbb{R}}

\newcommand{\Tr}{^\mathrm{T}}

\newcommand{\Xx}{\mathcal{X}}
\newcommand{\Yy}{\mathcal{Y}}
\newcommand{\Kk}{\mathcal{K}}

\newcommand{\bb}[1]{\bm{\mathrm{#1}}}

%% file: 1_intro.tex
\section{Introduction}

Estimating the correspondence between 3D shapes is among the fundamental problems in computer vision, geometry processing and graphics with a wide spectrum of applications ranging from 3D scene understanding to texture mapping and animation. 
%
Of particular interest is the case in which the objects are allowed to deform non-rigidly. In this setting, research has mainly focused on minimizing a measure of distortion between the input shapes, reaching in recent years very high levels of accuracy \cite{kaick2010survey}. However, point-wise accuracy often comes under restricting requirements (isometry assumption), or at the price of a lack of useful properties on the computed map, namely {\em bijectivity} (each point on either shape should have exactly one corresponding point on the other) and {\em smoothness} (nearby points should match to nearby points).

In this paper, we introduce a novel method to recover smooth bijective maps between deformable shapes. Contrarily to previous approaches, we do not rely on the assumption that the two shapes are isometric. We phrase our matching problem by using the language of statistical inference, whereas the input to our algorithm is either 1) a sparse collection of point-wise matches (as few as two) which are used as landmark constraints to recover the complete map, or 2) a dense, noisy, possibly non-surjective and non-smooth map which is converted to a better map with higher accuracy and the aforementioned properties.

\begin{figure}[t]
	\begin{center}
		\begin{overpic}
			[width=0.49\linewidth]{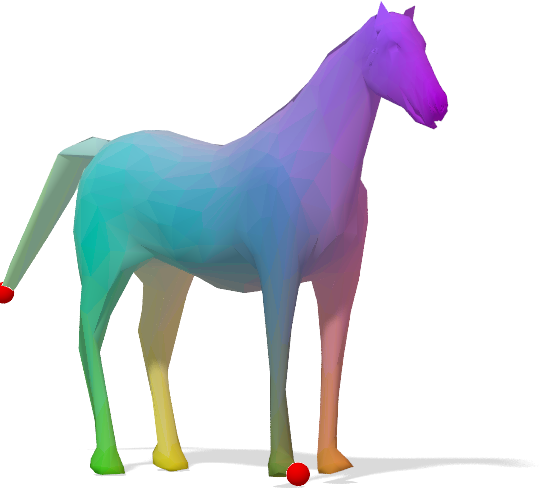}
		\end{overpic}
		\begin{overpic}
			[width=0.49\linewidth]{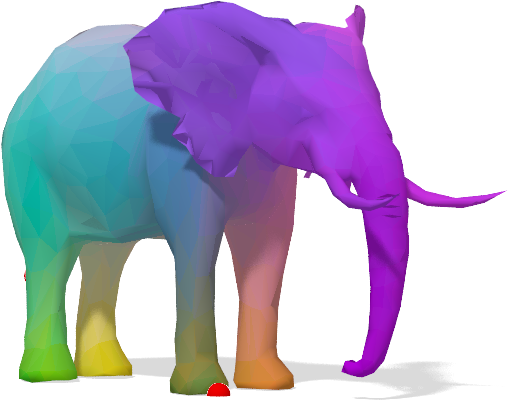}
		\end{overpic}
	\end{center}
	\caption{Our method can be used to recover a dense, smooth, bijective correspondence between highly non-isometric shapes from minimal input information. In this example, we initialize our algorithm with just two hand-picked matches (red spheres on tail and front leg). Correspondence quality is visualized by transferring colors from horse to elephant via the recovered map.\vspace{-3ex}}
	\label{fig:horseelephant}
\end{figure}

\subsection{Related works} 
\label{sec:related}
A traditional approach to correspondence problems is finding a {\em point-wise} matching between (a subset of) the points on two or more shapes. 
{\em Minimum-distortion methods} establish the matching by minimizing some structure distortion, which can include similarity of local features 
\cite{OvMe*10,BronsteinK10,WKS,ZaBo*09}, 
geodesic \cite{Memoli:2005,bro:bro:kim:PNAS,koltun} or diffusion distances \cite{lafon:05:LOCAL}, 
or a combination thereof \cite{torresani2008feature}. 

Typically, the computational complexity of such methods is high, and there have been several attempts to alleviate the computational complexity using hierarchical 
\cite{sahillioglu2012} 
or subsampling \cite{tevs2011intrinsic} methods. 
Several approaches formulate the correspondence problem as quadratic assignment and employ different relaxations thereof \cite{umeyama1988eigendecomposition,leordeanu2005spectral,rodola2012game,aflalo2015convex,koltun,lipman15}. 
Algorithms in this category typically produce guaranteed bijective correspondences between a sparse set of points, or a dense correspondence suffering from poor surjectivity.

{\em Embedding methods} try to exploit some assumption on the correspondence (e.g. approximate isometry) in order to parametrize the correspondence problem with a few degrees of freedom. Elad and Kimmel \cite{ela:kim:FLATTEN} used multi-dimensional scaling to embed the geodesic metric of the matched shapes into a low-dimensional Euclidean space, where alignment of the resulting ``canonical forms'' is then performed by simple rigid matching (ICP) \cite{ChenMedioni:91:ICP,bes:mck:SURFACEMATCH}. 
%
The works of \cite{Mateus08,shtern2013matching} used the eigenfunctions of the Laplace-Beltrami operator as embedding coordinates and performed matching in the eigenspace. 
Lipman \emph{et al.} \cite{Lipman2011,KimLCF10,kim2011blended} used conformal embeddings into disks and spheres to parametrize correspondences between homeomorphic surfaces as M{\"o}bius transformations. Despite their overall good performance, the majority of the matching procedures performed in the embedding space often produces noisy correspondences at fine scales, and suffers from poor surjectivity. 
More recently, in \cite{aigerman2014lifted,aigerman2016hyperbolic} the authors obtain a bijective correspondence by first computing compatible embeddings of the two shapes, and then aligning the embeddings through the use of sparse input correspondences.
As opposed to point-wise correspondence methods, {\em soft correspondence} approaches assign a point on one shape to more than one point on the other. 
Several methods formulated soft correspondence as a mass-transportation problem \cite{Me11,solomon2012soft}. 
Ovsjanikov \emph{et al.} \cite{ovsjanikov2012functional} introduced the {\em functional correspondence} framework, modeling the correspondence as a linear operator between spaces of functions on two shapes, which has an efficient representation in the Laplacian eigenbases. 
This approach was extended in several follow-up works \cite{pokrass2013sparse,kovnatsky15,SGMDS,rodola16-partial} .
A point-wise map is typically recovered from a low-rank approximation of the functional correspondence by a matching procedure in the representation basis, which also suffers from poor surjectivity. 
A third class of matching methods formulates the correspondence problem as an optimization problem in the \emph{product space} of the considered shapes. Windheuser {\em et al.} \cite{Windheusericcv11} seek for a two-dimensional minimal surface in the four-dimensional product space of the two input surfaces; this was later extended to a 2D-to-3D setting by L\"ahner \etal~\cite{lahner16}. Making use of the graph structure of the considered shapes, the discretization leads to an integer linear program on the product mesh where desirable properties of the matching such as smoothness and surjectivity become linear constraints. However, the computational complexity is prohibitive even for a modestly-sized problem. 
\subsection{Main contributions}
\begin{figure}
	\begin{center}
		\begin{overpic}[width=0.475\linewidth]{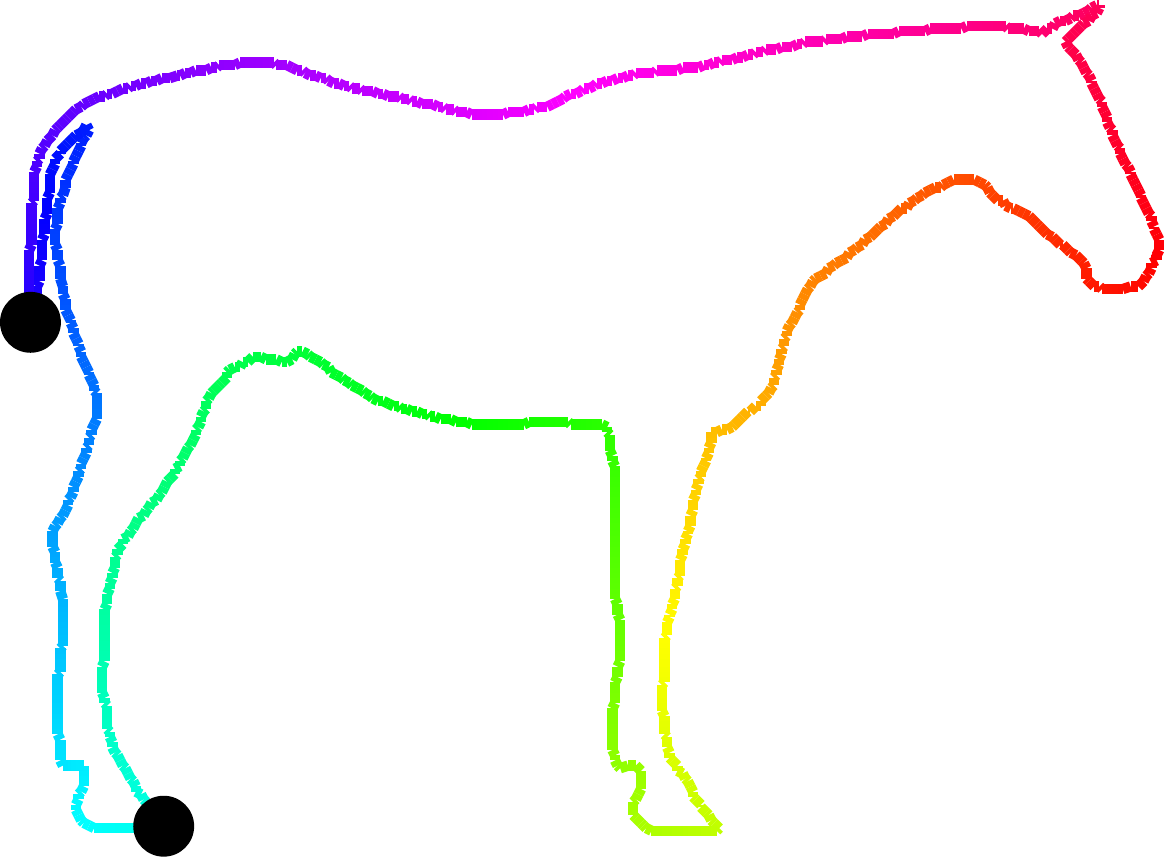}
		\end{overpic}
		\begin{overpic}[width=0.48\linewidth]{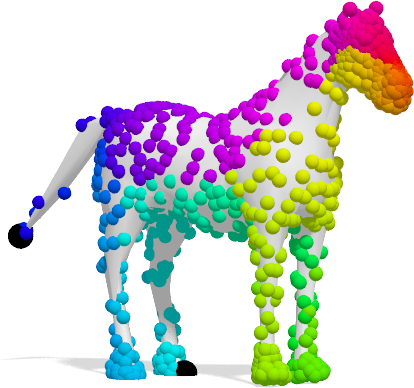}
		\end{overpic}
	\end{center}
	\vspace{-1ex}
	\caption{The Product Manifold Filter (PMF) can be applied to a variety of problems that are aiming for bijective, smooth mappings between metric spaces. Here we map a 2D shape (contour) to a 3D shape. We initialize the PMF with two semantically meaningful matches (black spheres) and obtain a dense semantically meaningful bijection.\vspace{-3ex}}
	\label{fig:2D3D}
\end{figure}

\vspace{-1ex}
Many of the works mentioned above provide a matching that is neither bijective nor smooth. In some cases the matching is only available as a sparse set of points in the product space of the two shapes.
We treat these matchings as corrupted versions of the latent correspondence and propose the \emph{Product Manifold Filter (PMF)}, a framework that increases the quality of the input mapping.
We show that the considered filter leads to a linear assignment problem (LAP) guaranteeing bijective correspondence between the shapes. Despite the common wisdom, we demonstrate that the problem is efficiently solvable for relatively densely sampled shapes by means of the well-established auction algorithm \cite{bertsekas} and a simple multi-scale approach. Unlike many of the previously mentioned techniques that assume the shapes to be (nearly) isometric, we allow them to undergo more general deformations (Figure \ref{fig:horseelephant}) or even have different dimensionality (Figure \ref{fig:2D3D}).

Finally, we present a significant amount of empirical evidence that the proposed smoothing procedure consistently improves the quality of the input correspondence coming from different algorithms, including point-wise recovery methods from functional map pipelines. We also show the performance of PMF as an interpolator of sparse input correspondences.

%% file: 2_method.tex
\vspace{-1ex}
\section{A probabilistic framework}
\vspace{-1ex}

We consider a pair of three-dimensional shapes that are represented by their boundaries $\Xx$ and $\Yy$, two-dimensional manifolds embedded in $\Rr^3$ and thus equipped with intrinsic metrics $d_\Xx$ and $d_\Yy$. Our goal is to find a semantically meaningful correspondence between $\Xx$ and $\Yy$. A correspondence is a \emph{diffeomorphism} $\pi: \Xx \rightarrow \Yy$, i.e., a smooth mapping with a smooth inverse. 
We do not make any other assumptions such as isometry.
The correspondence $\pi$ can be represented as a two-dimensional manifold $\Pi$ in the four-dimensional product space $\Xx \times \Yy$: a pair $(x,y)$ belongs to $\Pi$ iff $\pi(x) = y$. We henceforth assume that the true correspondence $\pi$ between $\Xx$ and $\Yy$ and the manifold $\Pi$ representing it are latent.

Let $\{(x_k,y_k)\}_{k\in \Kk} \subset \Pi$ be a possibly sparse sample of the said manifold. For example, these can be pairs of corresponding points on $\Xx$ and $\Yy$ computed using a feature detector followed by descriptor matching. In practice, we only have access to a noisy realization of these points, $\{(\xi_k,\eta_k)\}_{k\in \Kk}$, which we assume to admit a separable i.i.d. Gaussian density, $f(\xi_k,\eta_k) \propto K(d_\Xx(x_k,\xi_k)) K(d_\Yy(y_k,\eta_k))$, where
$$
K(d) = \exp\left(-\frac{d^2}{2\sigma^2} \right)
$$
is an unnormalized Gaussian kernel with the parameter $\sigma^2$. Note that the density on the manifolds is expressed in terms of the intrinsic metrics $d_\Xx$ and $d_\Yy$.

\begin{figure*}
	\begin{center}
		\begin{overpic}[width=0.75\linewidth]{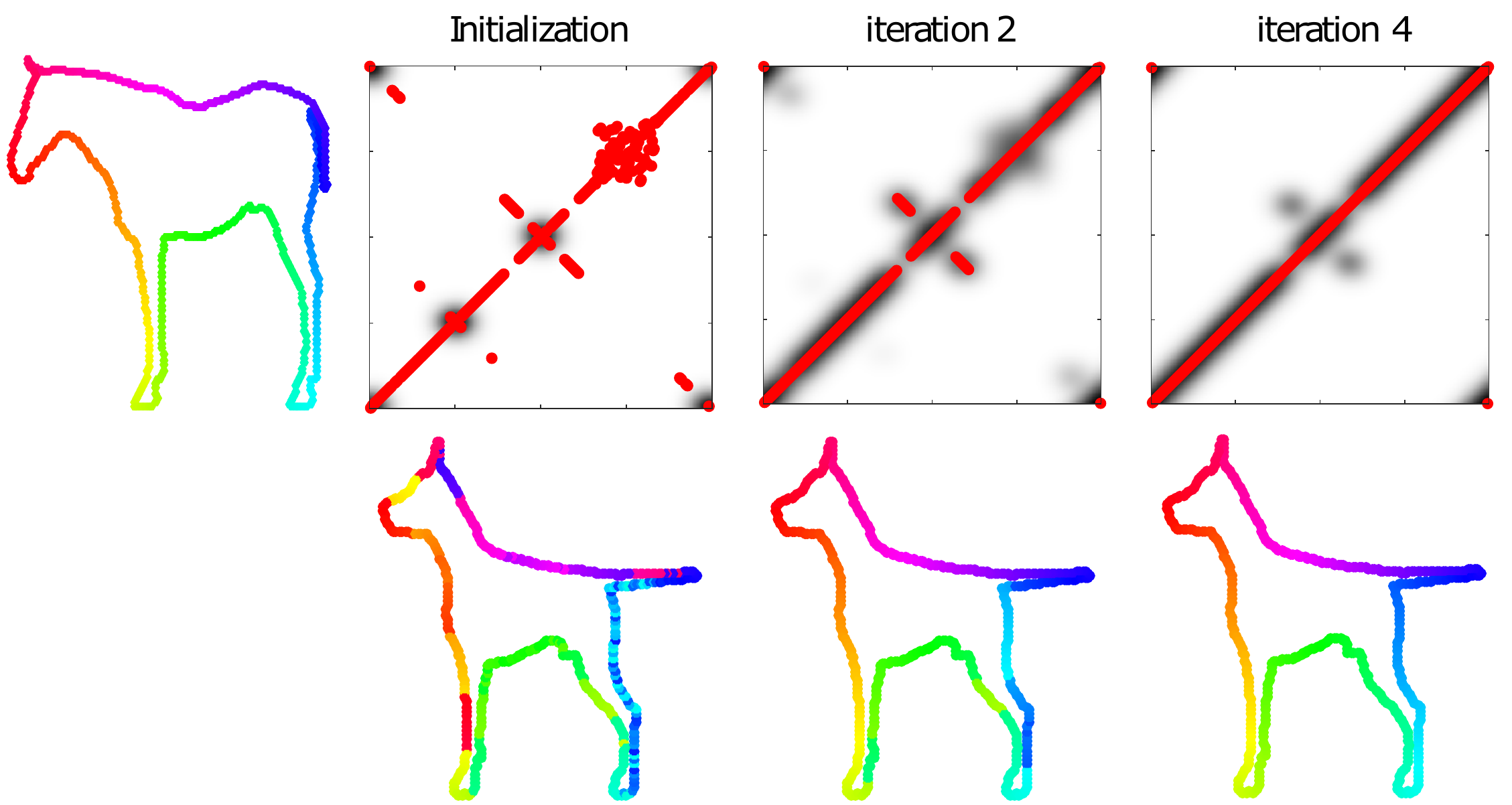}
		\end{overpic}
	\end{center}
	\vspace{-1ex}
	\caption{Conceptual illustration of our method on one-dimensional manifolds.
		Shown are iterations of PMF ($|\Kk|=3$ sparse matches as initialization).
		Top: Kernel density estimation $f(x,y)$ as defined in \eqref{eq:kernel} in the product space of the two shapes $\Xx$ and $\Yy$. Dark areas correspond to higher density.     
		According to \eqref{eq:est_bij}, consistently maximizing $f(x,\cdot)$ gives a bijective and smoothed matching (red curve in product space) which is used to derive the density estimate in the next iteration. 
		Bottom: matching visualized via color transfer. Shapes are parametrized counter-clockwise with the origin of the product space corresponding to the noses of horse and dog. Note the circular boundary conditions of the product space. 
		\vspace{-2ex}
	}
	\label{fig:2d}
\end{figure*}

Given the set of noisy corresponding points $\{(\xi_k,\eta_k)\}_{k\in \Kk}$ as the input, our goal is to produce a faithful estimate of the correspondence $\pi$. We propose to estimate the latent manifold $\Pi$ via kernel density estimation in the product space $\Xx \times \Yy$. To that end, we estimate the density function using the Parzen sum
\begin{eqnarray}
f(x,y) &\propto& \sum_{k\in \Kk} K( d_\Xx(x,\xi_k) ) \, K( d_\Yy(y,\eta_k) ).
\label{eq:kernel}
\end{eqnarray}
For every point $x \in \Xx$, an estimate of $\pi(x)$ is given by a point $y$ maximizing $f(x,y)$,
\begin{equation}
\hat{\pi}(x) = \mathrm{arg} \max_{y} f(x,y).
\label{eq:est}
\end{equation}
One can further impose bijectivity of $\hat{\pi} : \Xx \rightarrow \Yy$ as a constraint, obtaining the following estimator of the entire map
\begin{equation}
\hat{\pi} = \mathrm{arg} \max_{ \hat{\pi} : \Xx \overset{1:1}{\rightarrow} \Yy  } \int_{\Xx}  f(x,\hat{\pi}(x)) dx.
\label{eq:est_bij}
\end{equation}
The process can be iterated as shown in the one-dimensional illustration in Figure \ref{fig:2d}.

Procedures (\ref{eq:est}) or (\ref{eq:est_bij}) have an area reduction effect on the manifold $\Pi$ producing a more regular version thereof and thus a more regular correspondence $\pi$. We interpret  (\ref{eq:est_bij}) as a filter of correspondences and will henceforth refer to it a \emph{product manifold filter} (PMF).
While we defer the rigorous proof of the area reduction  property to the extended version of the paper, in what follows, we illustrate it by a simple one-dimensional example.

\vspace{1ex}\noindent\textbf{One dimensional illustration.}
Let us consider a configuration of three points $\{x_-,x,x_+\}$ and the corresponding noisy points $\{y_-,y,y_+\}$ on a pair of one-dimensional manifolds $\Xx$ and $\Yy$ like those depicted in Figure \ref{fig:2d}. We assume that the points are directly given in arclength parametrization, such that $d_\Xx(x,x_\pm) = |x-x_\pm| = b$,  
$d_\Yy(y,y_-) = | y-y_- | = a$, and $d_\Yy(y,y_+) = |y-y_+| = a + \delta$. For convenience, we henceforth denote $x=y=0$, $x_\pm = \pm b$, $y_- = -a$ and $y_+ = a+\delta$. In this setting, the one-dimensional manifold $\Pi_0$ representing the input correspondence in the product space comprises two segments connecting $(-b,-a)$, $(0,0)$, and $(b,a+\delta)$, and its length is given by
$
L(\Pi_0) = \sqrt{b^2 + a^2} + \sqrt{b^2 + (a+\delta)^2}.
$

PMF maximizes the density function
\begin{eqnarray}
h(\hat{y}) &=& f(0,\hat{y}) = K(0) K(\hat{y}) + \\
&& K(b) K(\hat{y}+a) + K(b) K(\hat{y}-a-\delta) \nonumber\\
&=& K(\hat{y}) + K(b) ( K(\hat{y}+a) + K(\hat{y}-a-\delta) )\nonumber
\end{eqnarray}
over the values $\hat{y}$ for the point $y$.
First, we observe that since $K(b) > 0$, the global maximum of $h(\hat{y})$ has to be around $\hat{y} = 0$. For $\hat{y} = 0$ and $\delta = 0$, one has
$$
\frac{dh}{d\hat{y}} = K'(0) + K(b) ( K'(a) + K'(-a) )
$$
and
$$
\frac{d^2h}{d\hat{y}^2} = K''(0) + K(b) ( K''(a) + K''(-a) )
$$
Since $K'(0) = 0$ and $K'(-a) = -K'(a)$, the first derivative vanishes, while the fact that $K''(0) < 0$ and $K''(-a) = K''(a)$ implies that $\hat{y}=0$ is the maximum of $h$. 

Next, we perform perturbation analysis of the above maximizer by invoking the first-order Taylor expansion of $h$ around $(\delta, \hat{y}) = (0,0)$:
$$
\frac{\partial h}{\partial\hat{y}} \approx \left. \frac{\partial h}{\partial\hat{y}}\right|_{\hat{y} = 0,\delta = 0} + \hat{y}\left.\frac{\partial^2 h}{\partial\hat{y}^2}\right|_{\hat{y} = 0,\delta = 0}  + \delta\left.\frac{\partial^2 h}{\partial \hat{y} \partial \delta}\right|_{\hat{y} = 0,\delta = 0}.
$$
Demanding equality to zero yields the maximizer of the perturbed problem
\begin{eqnarray*}
\hat{y} & \approx &
\frac{K(b) K''(a) \delta }{ 2K(b) K''(a) + K''(0)  } 
=  \frac{\delta }{ 2 + \frac{K''(0)}{ K(b) K''(a)  }   } = c\delta.
\end{eqnarray*}
For $a < \frac{\sigma}{\sqrt{2}}$ the ratio in the denominator is positive and consequently $c\in (0,\frac{1}{2})$.

The length of the estimated manifold $\hat{\Pi}$ can be obtained using a series of first-order Taylor approximations,
\begin{eqnarray}
L(\hat{\Pi}) &=& \sqrt{b^2 + (a+c\delta)^2} + \sqrt{b^2 + (a+\delta-c\delta)^2} \nonumber\\
& \approx & L(\Pi_0) + \frac{a c\delta}{\sqrt{b^2 + a^2}} 
- \frac{(a+\delta) c\delta}{\sqrt{b^2 + (a+\delta)^2}} 
\nonumber\\
&\approx & L(\Pi_0) - \frac{cb^2}{(b^2+a^2)^{3/2}} \, \delta^2 < L(\Pi_0),
\end{eqnarray}
which manifests the length reducing effect of the PMF.

\subsection{Discretization}\label{sec:discrete}
\vspace{-1ex}

In what follows, we consider a discretization of problem  \eqref{eq:est_bij}. We assume the shape $\Xx$ to be discretized at $n$ points $\{x_i\}_{i=1}^n$
and the pairwise geodesic distances are stored in the matrix $\bb{D}_\Xx\in \Rr^{n\times n}$. Similarly, the shape $\Yy$ is discretized as $\{y_i\}_{i=1}^n$ and its pairwise distance matrix is denoted by $\bb{D}_\Yy\in \Rr^{n\times n}$. Given a (possibly sparse) collection of input correspondences $\{(\xi_k,\eta_k)\}_{k=1}^m$ the unnormalized kernel density estimation can be written as an $n \times n$ matrix
\begin{eqnarray}
\label{eq:f_matrix}
 \bb{F} &=& \bb{K}_\Xx\bb{K}_\Yy^T
\end{eqnarray}
with the matrices $\bb{K}_\Xx\in \Rr^{n\times m}$ and $\bb{K}_\Yy\in \Rr^{n\times m}$ given by
\begin{eqnarray}
 (\bb{K}_\Xx)_{ik} &=& K(d_\Xx(x_i,\xi_k)) \\
 (\bb{K}_\Yy)_{ik} &=& K(d_\Yy(y_i,\eta_k)).
\end{eqnarray}
%
%
%
%
%
The objective in \eqref{eq:est_bij} thus becomes

\begin{eqnarray}
 \int_\Xx f(x,\pi(x)) dx &=&  \int_{\Xx \times \Yy} f(x,y) \delta_{\pi(x)}(y) dy dx \nonumber\\
  &\approx &  \sum_{i,j=1}^n \bb{F}_{ij} \bb{P_{ji}} = \langle \bb{P},\bb{F}\rangle
\end{eqnarray}

with $\bb{P}\in \{0,1\}^{n\times n}$ being a permutation matrix representing a bijection between $\{x_i\}_{i=1}^n$ and $\{y_i\}_{i=1}^n$.
At some points it will be convenient to use the vector representation $p\in\{1,\ldots,n\}^n$ of $\bb{P}$.
Estimating the bijective correspondence as in \eqref{eq:est_bij} thus turns out to be a linear assignment problem (LAP) of the form


\begin{equation}
\hat{\bb{P}} = \mathrm{arg} \max_{ \bb{P}}  \langle \bb{P},\bb{F} \rangle\label{eq:LAP}
\end{equation}
where the optimization is performed over the space of all $n \times n$ permutation matrices. 

\subsection{Multiscale}
\label{sec:multi}

While linear assignment problems like \eqref{eq:LAP} can be solved in polynomial time, the memory consumption is quadratic
in the vertex set size $n$.
To alleviate this burden, we propose a multi-scale technique based on the assumption of local regularity of the manifold $\Pi$. 

Given two shapes discretized at $n$ points each, we perform farthest point sampling to obtain a hierarchy of $p$ multiscale representations consisting of $n_1<n_2<\ldots<n_p=n$ points. Each of the samplings comes with a sequence of sampling radii, $r^\Xx_1>r^\Xx_2>\ldots>r^\Xx_p$ and $r^\Yy_1>r^\Yy_2>\ldots>r^\Yy_p$, respectively.

For sufficiently large shapes, the $n\times n$ pairwise distance matrices $\bb{D}_\Xx$ and $\bb{D}_\Yy$ can be no more stored entirely in memory.
We follow \cite{aflalo2013spectral,litman2016spectrometer} and store only the projection of the latter matrices on the first $r$ eigenfunctions of the Laplacian resulting in an $n \times r$ matrix. 
The original distances are reconstructed on-demand, with negligible error as shown in \cite{aflalo2015optimality,litman2016spectrometer}.

We recursively apply a variant of the PMF to the sparse set of input matches obtained by the coarser scale:

\begin{eqnarray}
\label{eq:multi_lap}
\bb{P}_{i+1} &=& \mathrm{arg} \max_{\bb{P}\in \{0,1\}^{n_{i+1}^2}} \langle \bb{P},\bb{F}_i\ \rangle
\end{eqnarray}

where
\begin{equation}
\bb{F}_i(s,t) =  \bb{W}(s,t) \sum_{k=1}^{n_i}  \bb{K}_\Xx(s,k)\bb{K}_\Yy(t,p_i(k))\,.
\end{equation}
%
The weighting matrix $\bb{W}\in \{0,1\}^{(n_{i+1}^2\times n_i^2)}$ assures that the image of a point $x_s$ being in the vicinity of $x^i_k$ is constrained to be mapped to a point in the vicinity of $p(k)$ and vice versa (i.e., the matching and its inverse are supposed to be smooth):
\begin{eqnarray}
\lefteqn{ \bb{W}(s,t) =} \label{eq:cases} \\ 
&& \begin{cases}
  0 & \text{if } \exists k:  \bb{D}_\Xx(s,k) < r^\Xx_i \text{ and } \bb{D}_\Yy(t,p_i(k)) > 2  r^\Yy_i\\
  0 & \text{if } \exists k:  \bb{D}_\Yy(t,p_i(k)) < r^\Yy_i \text{ and } \bb{D}_\Xx(s,k) > 2  r^\Xx_i\\
  1 & \text{otherwise}
 \end{cases}\nonumber
 \end{eqnarray}

This construction leads to a sparse payoff matrix corresponding to a smaller space of feasible permutations, so that the corresponding LAP can be solved efficiently. 
Note the factor $2$ in (\ref{eq:cases}). 
Since we cannot guarantee the Voronoi cells on the two shapes to have the same number of points and we want to be able to remove errors from the coarser scale, we permit moving a point to an adjacent Voronoi cell.

%% file: 4_results.tex
\begin{figure}[tb]
	\begin{center}
		\begin{overpic}[width=0.3\linewidth]{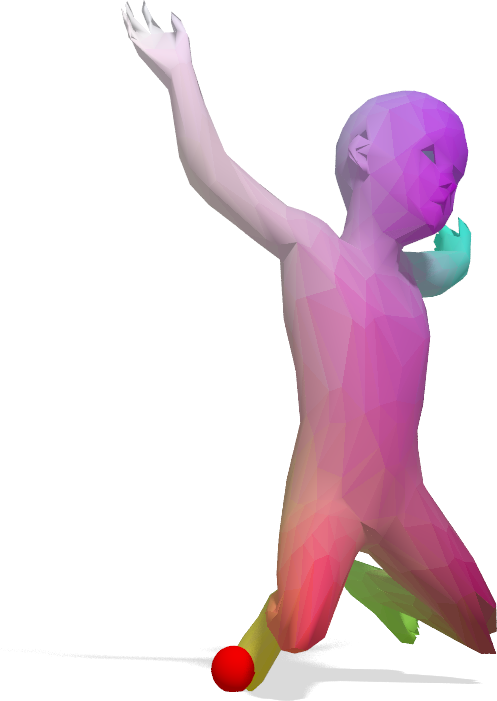}
		\end{overpic}\hspace{0.2cm}
		\begin{overpic}[width=0.33\linewidth]{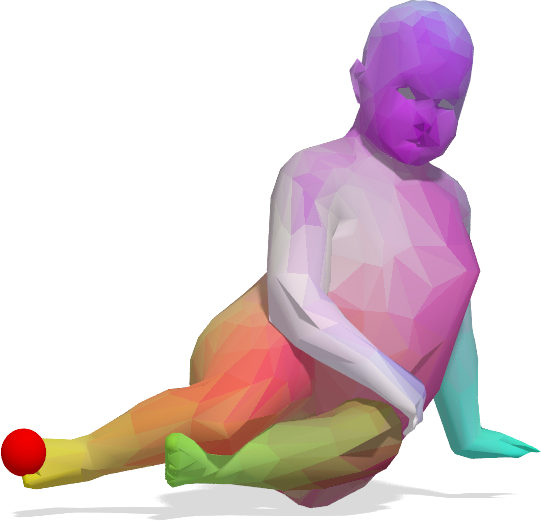}
		\end{overpic}
		\begin{overpic}[width=0.32\linewidth]{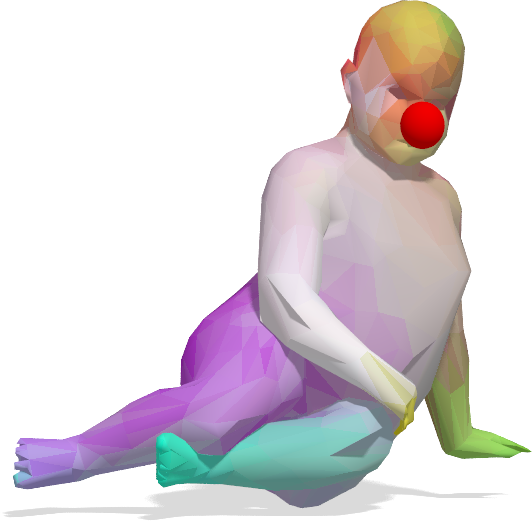}
		\end{overpic}
	\end{center}
    \vspace{-1ex}
	\caption{\label{fig:onepoint}Our method finds smooth bijective maps between non-isometric shapes even when one single match is given as input (marked as small red spheres). Note that the map remains smooth even if the initial match is wrong (rightmost column).}
\end{figure}

\begin{figure*}[t!]
	\begin{center}
		\begin{overpic}[width=0.15\linewidth]{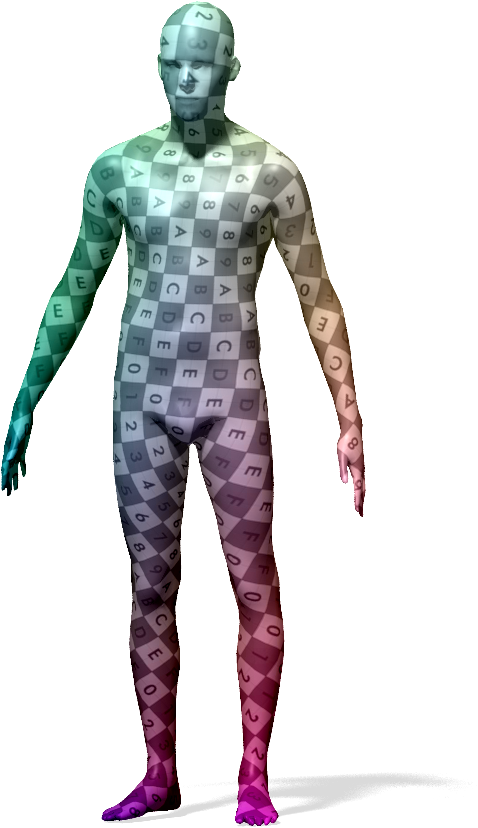}
        \put(12,-8){\footnotesize Reference}
		\end{overpic}
		\begin{overpic}[width=0.15\linewidth]{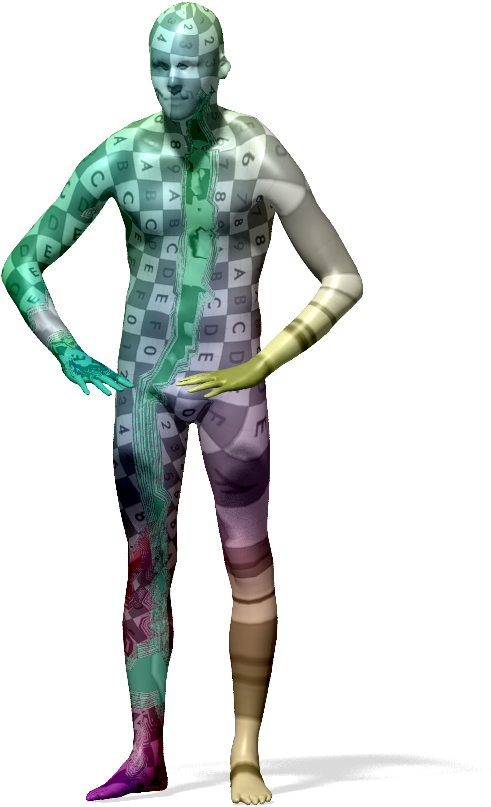}
         \put(19,-8){\footnotesize \cite{aigerman2016hyperbolic}}
		\end{overpic}
		\hspace{-0.6cm}\begin{overpic}[width=0.15\linewidth]{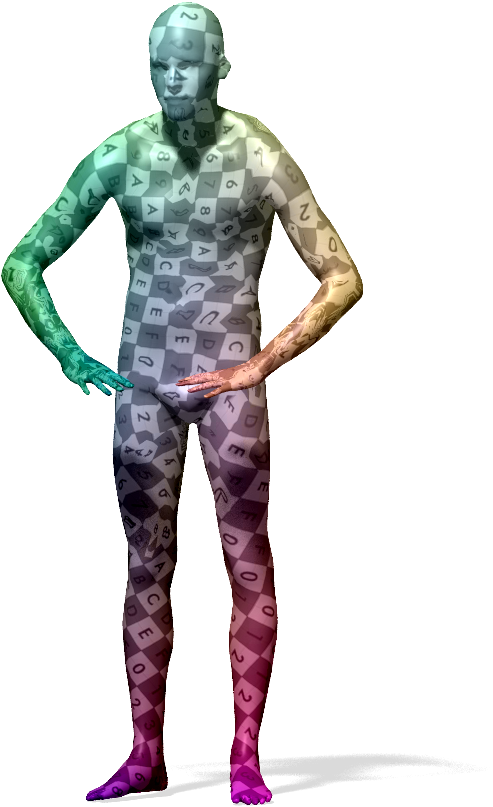}
         \put(-3.5,94){\footnotesize 3}
         \put(17,-8){\footnotesize Ours}
		\end{overpic}
		\begin{overpic}[width=0.15\linewidth]{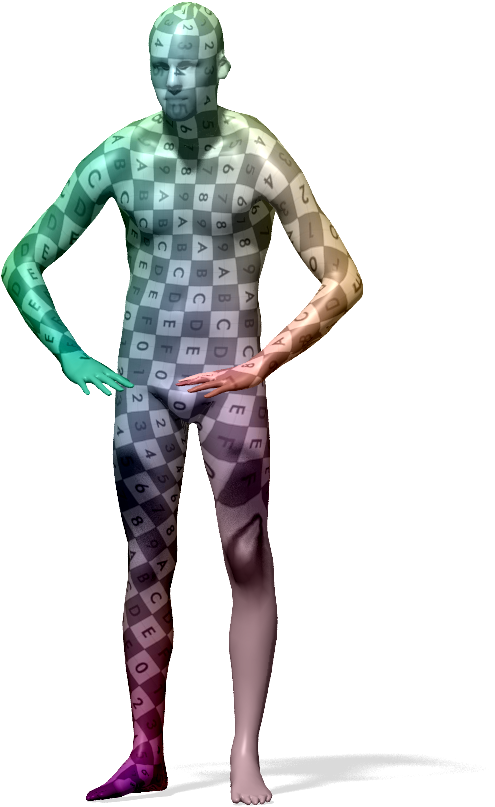}
         \put(19,-8){\footnotesize \cite{aigerman2016hyperbolic}}
		\end{overpic}
		\hspace{-0.6cm}\begin{overpic}[width=0.15\linewidth]{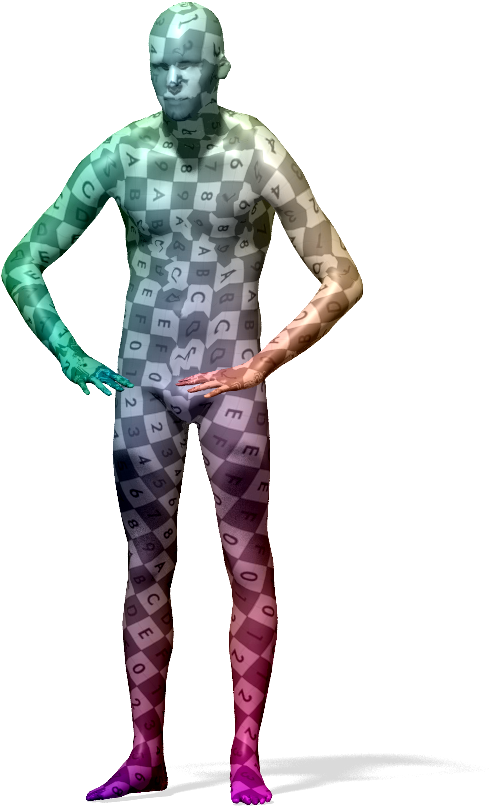}
         \put(-3.5,94){\footnotesize 5}
         \put(17,-8){\footnotesize Ours}
		\end{overpic}
		\begin{overpic}[width=0.15\linewidth]{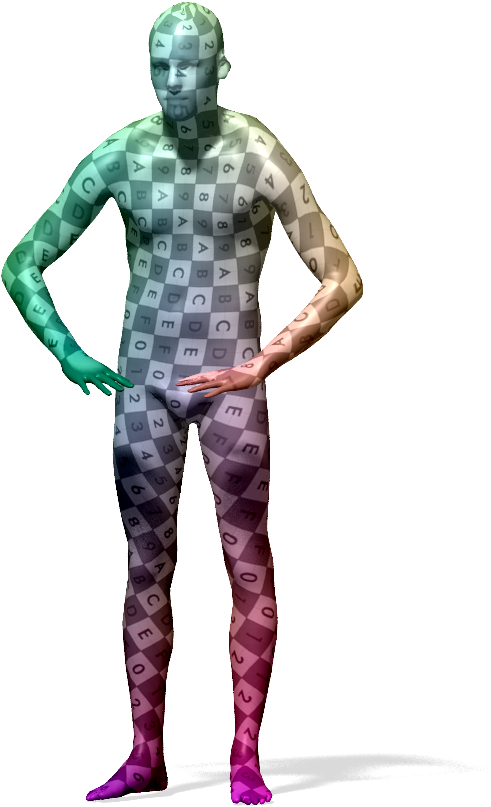}
         \put(19,-8){\footnotesize \cite{aigerman2016hyperbolic}}
		\end{overpic}
		\hspace{-0.6cm}\begin{overpic}[width=0.15\linewidth]{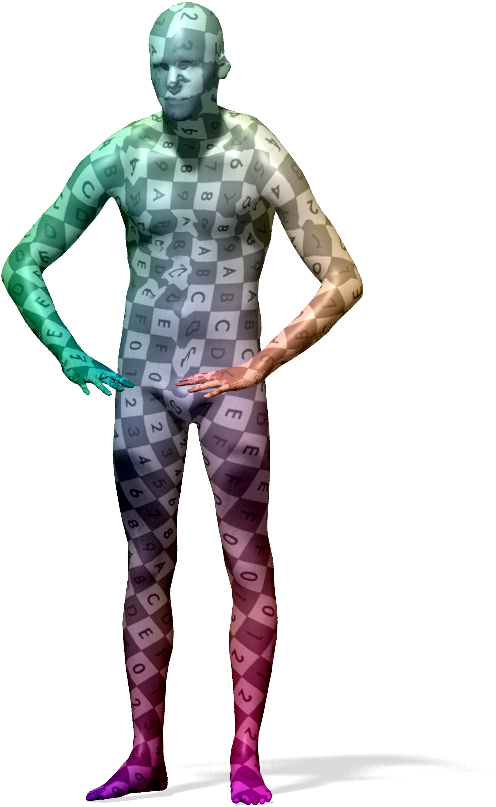}
         \put(-4.5,94){\footnotesize 10}
         \put(17,-8){\footnotesize Ours}
		\end{overpic}
	\end{center}
	\caption{\label{fig:lipman}Comparison between our method and the  method of \cite{aigerman2016hyperbolic} at increasing number of input matches (reported on top). Both methods produce smooth, guaranteed bijective solutions; our method requires little computational effort (a few minutes as opposed to $\sim$1 hour for \cite{aigerman2016hyperbolic}), and yields in comparison more accurate solutions when fed with a very sparse input.\vspace{-2ex}}
\end{figure*}
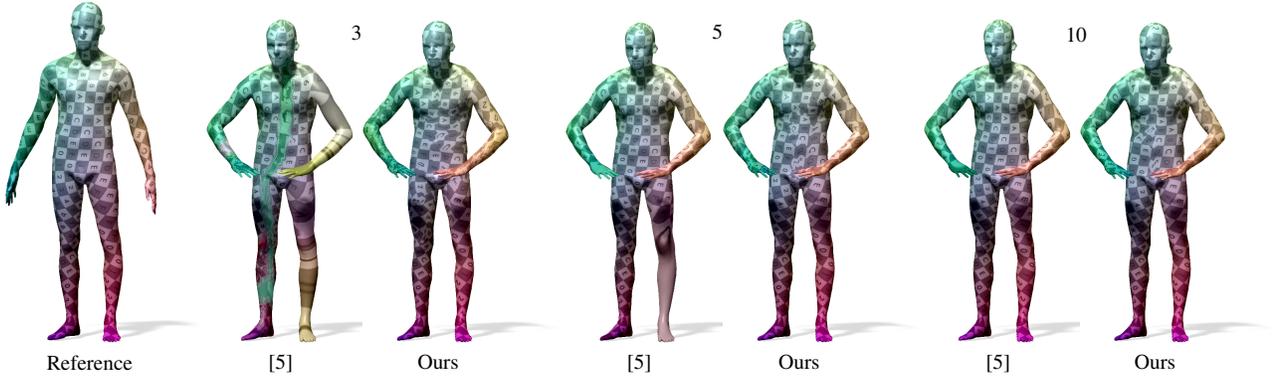

\section{Experiments}
\label{sec:experiments}

While our method can be applied to a variety of problems aiming at bijective and smooth mappings between metric spaces (see Figure~\ref{fig:2D3D} for an extreme case), here we focus on the recovery of a correspondence between non-rigid and possibly non-isometric 3D shapes. We show the performance of our method in two very different scenarios, namely refinement of noisy dense correspondence, and completion of sparse correspondence. 
%
%
We additionally demonstrate the performance of our multi-scale technique by recovering bijective correspondences between high resolution shapes.

\subsection{Recovery from sparse correspondences}
\vspace{-1ex}
In our first set of experiments we consider a scenario in which the input shapes come with a (possibly very sparse) collection of initial matches. These, in turn, can be obtained by a sparse non-rigid matching technique such as \cite{rodola2012game} or be hand-picked, depending on the application. In these experiments we compare PMF with the Tutte embedding approach recently introduced in ~\cite{aigerman2016hyperbolic}. Similarly to PMF, this approach produces guaranteed bijective and smooth maps starting from a sparse set of point-wise matches; to our knowledge, this method represents the state-of-the-art for this class of problems.

The results of this comparison are shown in Figure~\ref{fig:lipman}. The input matches were obtained by mapping farthest point samples on a reference shape via the ground-truth correspondence to the target shape, and are visualized by transferring a texture from reference to target via the recovered {\em dense} map. As we can read from the plots, our approach yields maps of better quality when fewer than ten matches are provided as the input, and maps of comparable quality when more matches are available. It is important to note that while our method still produces meaningful solutions when just one or two matches are given as the input (see Figures~\ref{fig:horseelephant}, \ref{fig:onepoint}), the approach of \cite{aigerman2016hyperbolic} has the theoretical minimum of five matches; furthermore, the latter approach gives different solutions depending on the specific ordering of the inputs, while our method is invariant to their permutations. Finally, as we demonstrate in the next section, a key ability of our method is being able to recover correct maps from noisy inputs, while the Tutte approach requires exact input.

\subsection{Recovery from noisy input}

In this set of experiments we assume to be given a low-rank approximation of the latent correspondence $\bb{P}$ in terms of a functional map 
\begin{eqnarray}
 \bb{C} &=& \bb{\Psi}^T \bb{P} \bb{\Phi} \in \Rr^{r\times r},
\end{eqnarray}
where $\bb{\Phi},\bb{\Psi}\in \Rr^{n\times r}$ are truncated orthonormal bases on $\Xx$ and $\Yy$. We refer the reader to the original paper \cite{ovsjanikov2012functional} for details and allow ourselves to condense its ideas to the above equation. 

While a plurality of methods for finding $\bb{C}$ have been proposed in the last years, there currently exist only three approaches to recover a point-wise correspondence matrix $\bb{P}$ from it.
In \cite{ovsjanikov2012functional} the authors proposed to recover a pointwise correspondence between $\Xx$ and $\Yy$ by solving the nearest-neighbor problem (NN)
\begin{align}\label{eq:nn1}
\min_{\bb{P} \in \{0,1\}^{n \times n}}~& \|\bb{C}\bb{\Phi}\Tr - \bb{\Psi}\Tr \bb{P}\|_ \mathrm{F}^2 & 
\mathrm{s.t.}~&\bb{P}\Tr\bb{1}=\bb{1}\,.
\end{align}
alternated with an orthogonality-enforcing refinement of $\bb{C}$ (ICP).
A variant is its bijective version (Bij. NN)

\begin{align}\label{eq:lap}
\min_{\bb{P} \in \{0,1\}^{n \times n}}~& \|\bb{C}\bb{\Phi}\Tr - \bb{\Psi}\Tr \bb{P}\|_ \mathrm{F}^2 &
\mathrm{s.t.}~&\bb{P}\Tr\bb{1}=\bb{1}\,,~\bb{P}\bb{1}=\bb{1}\,.
\end{align}

The orthogonal refinement of \eqref{eq:nn1} assumes the underlying map to be area-preserving \cite{ovsjanikov2012functional}, and is therefore bound to fail in case the two shapes are non-isometric.
Rodol\`{a} \emph{et al.} \cite{rodola-vmv15} proposed to consider the non-rigid counterpart for a given $\bb{C}$:
\begin{align}
\label{eq:cpd}
\min_{\bb{P}\in [0,1]^{n\times n}}~& D_\mathrm{KL}(\bb{C}\bb{\Phi}\Tr,\bb{\Psi}\Tr\bb{P})
 + \lambda \| \bb \Omega (\bb{C}\bb{\Phi}\Tr - \bb{\Psi}\Tr\bb{P}) \|^2\nonumber\\
\mathrm{s.t.}~ &\bb{P}\Tr\bb{1}=\bb{1}\,.
\end{align}
Here $D_\mathrm{KL}$ denotes the Kullback-Leibler divergence between probability distributions, $\bb\Omega$ is a low-pass operator promoting smooth velocity vectors, and $\lambda>0$ controls the regularity of the assignment. The problem is then solved via expectation-maximization by the coherent point drift algorithm (CPD) \cite{myronenko10}.

We construct the low-rank functional map using the known ground-truth correspondences between the shapes. Since this is supposed to be the ideal input for all the competing methods, we abandon the refinement step in \eqref{eq:nn1}.
Correspondences returned by the other methods are treated as noisy realizations of the latent bijection and are recovered via PMF with $\sigma^2$ set to $2\%$ of the target shape area. 

\begin{figure}[t]
	\begin{center}
		\begin{overpic}[width=.97\linewidth]{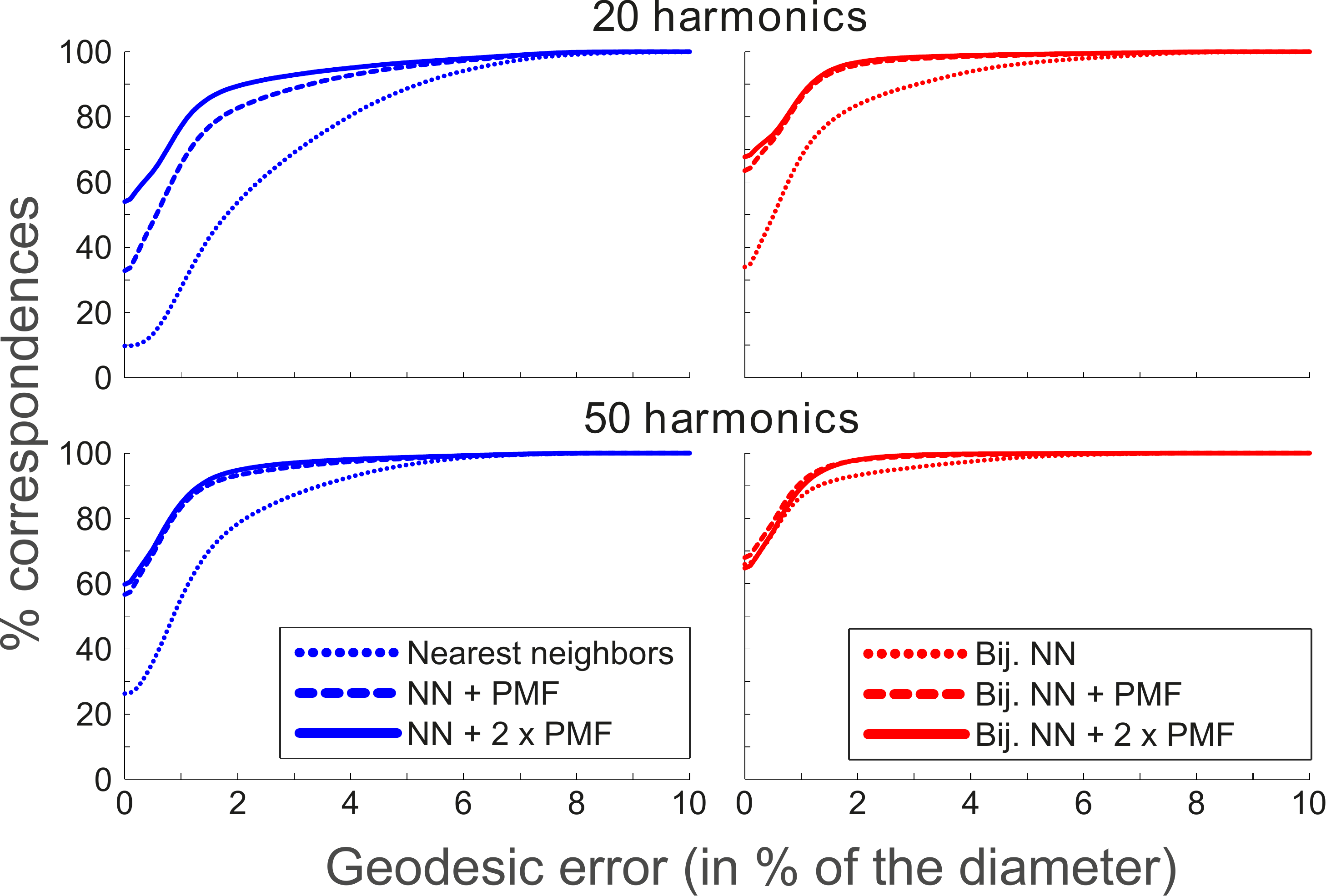}
		\end{overpic}
	\end{center}
	\caption{Quantitative comparison of methods for pointwise correspondence on the non-isometric FAUST dataset (about 7K vertices).}
	\label{fig:funmap_faust}
\end{figure}

\begin{figure*}
	\begin{center}
		\begin{overpic}[width=.98\linewidth]{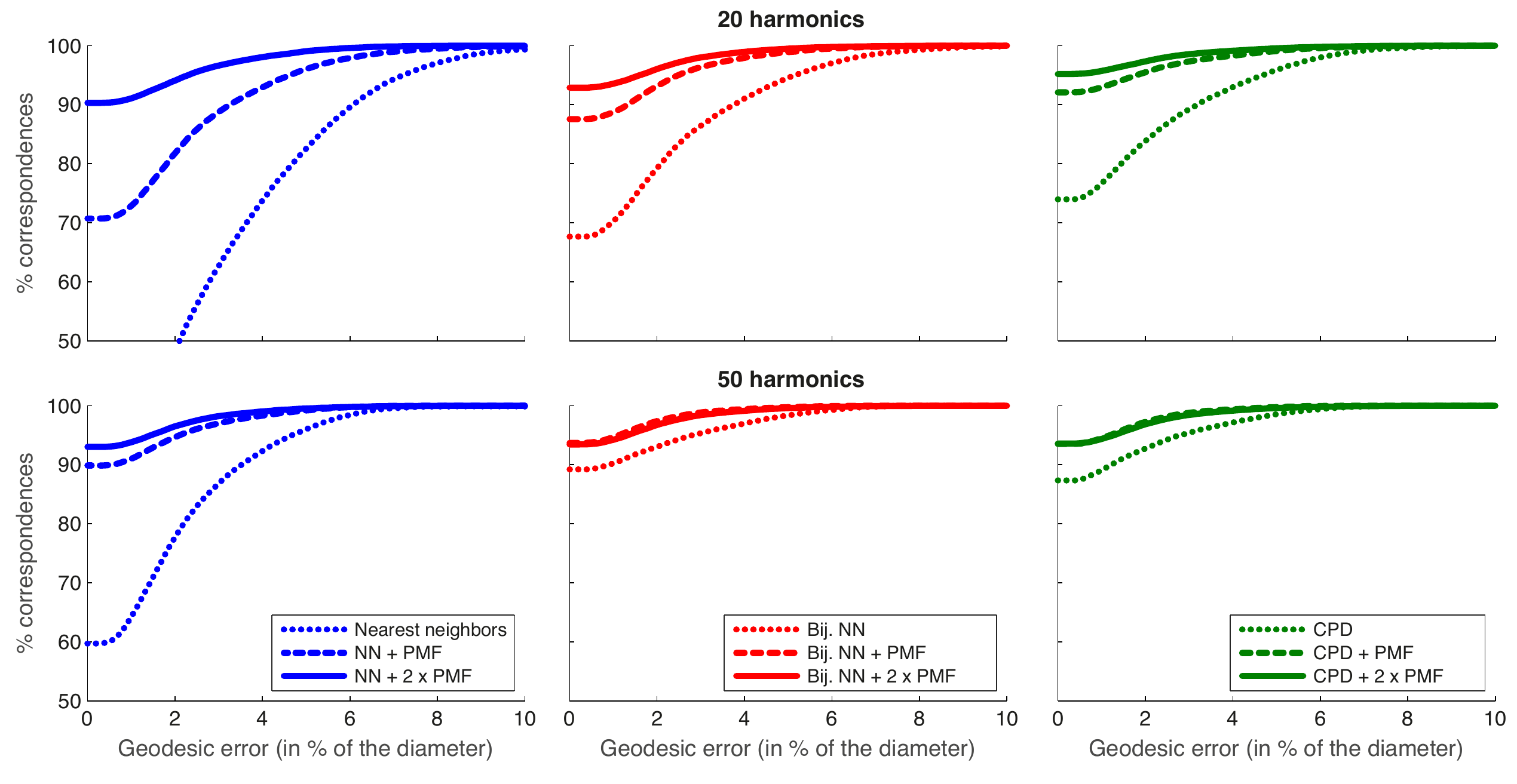}
		\end{overpic}
	\end{center}
	\caption{Quantitative comparison of methods for pointwise correspondence recovery from a functional map ($20$ and $50$ eigenfunctions). We matched 70 pairs from the near-isometric SCAPE dataset (1K). Plotted are the histograms of geodesic errors. Filtering the results of nearest neighbors (left) outperforms the state of the art method (right) while having only a fraction of its runtime (Table \ref{tab:run_times}). Even better results are achieved under affordable runtimes when initializing the PMF estimator with the result of bijective NN (center).}
	\label{fig:funmap_scape}
\end{figure*}

\begin{figure*}
	\begin{center}
		\begin{overpic}[width=1\linewidth]{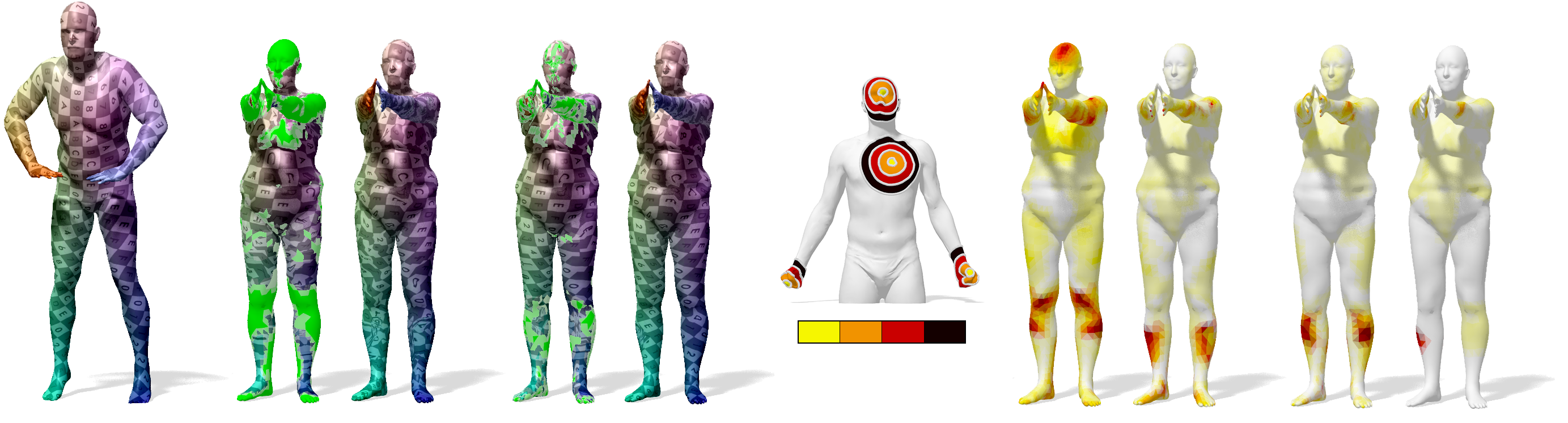}
			\put(51.5,3.5){\tiny $1\%$}
			\put(54.5,3.5){\tiny $3\%$}
			\put(57.5,3.5){\tiny $5\%$}
			\put(60.5,3.5){\tiny $7\%$}
			\put(54.5,2.25){\tiny $\times \mathrm{diam}$}
			\put(17,25.5){\small NN}
			\put(21.5,25.5){\small NN+PMF}
			\put(34,25.5){\small CPD}
			\put(39,25.5){\small CPD+PMF}
			\put(67,25.5){\small NN}
			\put(71,25.5){\small NN+PMF}
			\put(83.5,25.5){\small CPD}
			\put(88.5,25.5){\small CPD+PMF}
		\end{overpic}
	\end{center}
	\caption{Qualitative comparison of methods for pointwise correspondence recovery from a functional map. Current methods such as nearest neighbors (NN) and coherent point drift (CPD) suffer from bad accuracy and lack of surjectivity. Applying the proposed estimation to either of them gives a guaranteed bijective matching with high accuracy and improved smoothness. Left: We visualize the accuracy of the methods by transferring texture from the source shape $\Xx$ to the target shape $\Yy$. Neither NN nor CPD produce bijective mappings. The lack of surjectivity is visualized by assigning a fixed color (green) to not-hit points. Right: The geodesic error (distance between ground-truth and recovered match, relative to the shape diameter) induced by the matching is visualized on the target shape $\Yy$.}
	\label{fig:funmap_qual}
\end{figure*}
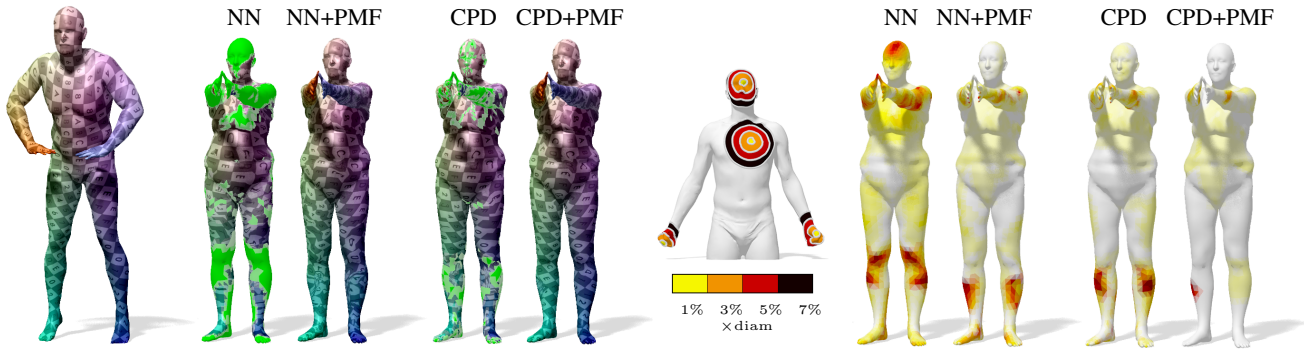

We show quantitative comparisons on 71 pairs from the SCAPE dataset \cite{SCAPE} (near isometric, 1K vertices) and 100 pairs from the FAUST dataset \cite{FAUST} (including inter-class pairs, 7K vertices). In Figures  \ref{fig:funmap_scape} and  \ref{fig:funmap_faust} we compare the correspondence accuracy, while in Figure \ref{fig:funmap_qual} we visualize how lack of smoothness, bijectivity and accuracy affect texture transfer.

The accuracy of all input matchings is increased by applying the product manifold filter. To our knowledge, the matchings obtained by the PMF are the most accurate ones that can be recovered from this type of low-rank approximation.
While linear assignment problems are known to be time demanding to solve for larger numbers of variables, the most dramatic increase of run time occurs when applying the coherent point drift algorithm (see Table \ref{tab:run_times}). 

\begin{figure}
	\begin{center}
		\begin{overpic}[width=0.5\linewidth]{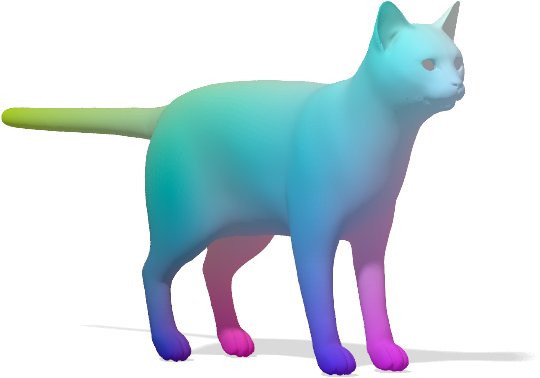}
		\end{overpic}\hspace{0.3cm}
		\begin{overpic}[width=0.4\linewidth]{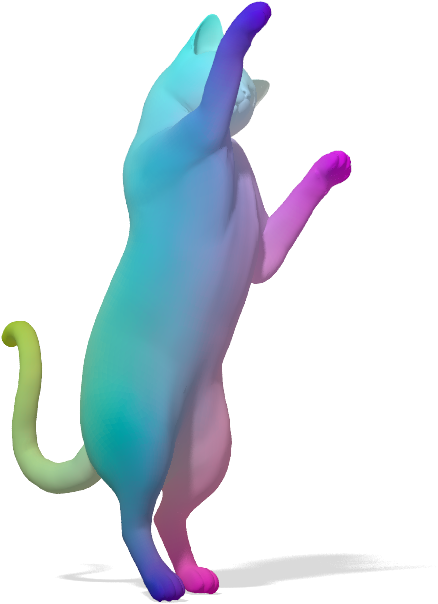}
		\end{overpic}
	\end{center}
    \vspace{-1ex}
	\caption{
		Result of our method on two cat shapes from TOSCA \cite{Bronstein:2008}. 
		This high resolution shape has $27894$ vertices, making it infeasible to store the entire pairwise distance matrix in memory. 
		Our multi-scale approach recovered a smooth matching from only $20$ sparse correspondences given as the input using five hierarchical scales as detailed in Section~\ref{sec:multi}.		
	}
	\label{fig:cats_multi}
\end{figure}

\begin{figure}[h]
	\begin{center}
		\begin{overpic}[width=0.22\linewidth]{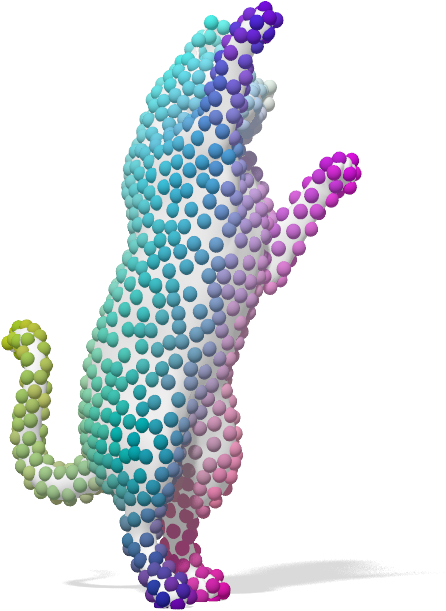}
		\end{overpic}
        \begin{overpic}[width=0.22\linewidth]{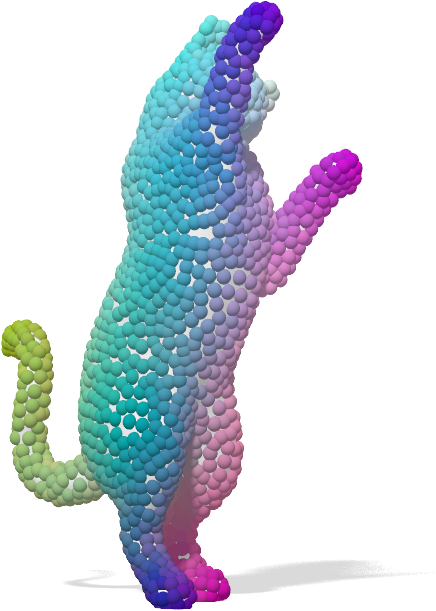}
		\end{overpic}
		\begin{overpic}[width=0.22\linewidth]{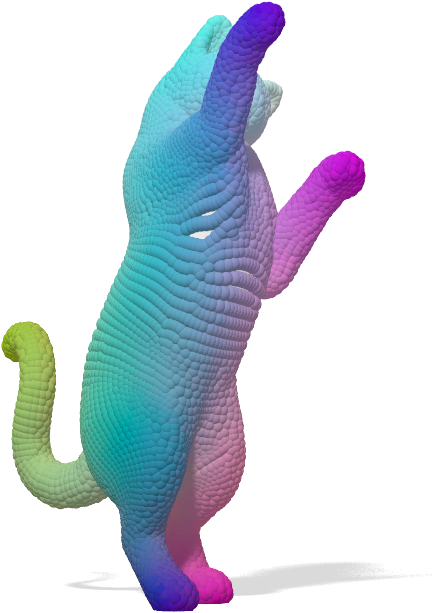}
		\end{overpic}
		\begin{overpic}[width=0.22\linewidth]{images/N_cat.png}
		\end{overpic}
	\end{center}
    \caption{Visualization of the multiscale iterations evaluated in Fig.~ \ref{fig:cat_err_curve}. From left to right: 1K, 2K, 8K, 28K (all) vertices.}
    \label{fig:cats_iters}
\end{figure}

\subsection{Recovering high-resolution correspondences using multiscale}

\begin{figure}	
	\includegraphics[width=1\linewidth]{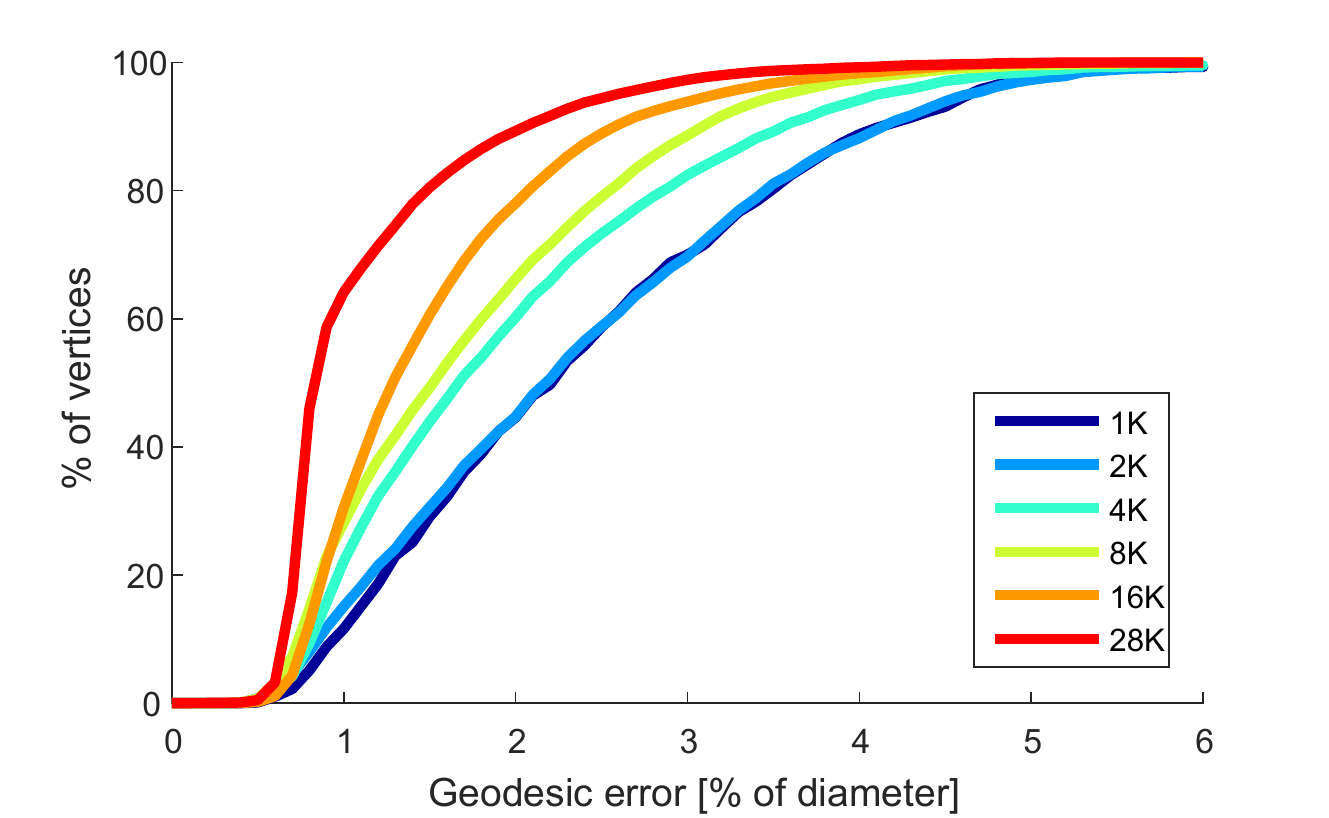}
	\caption{Quantitative analysis of correspondences between the two cats shown in Figure~\ref{fig:cats_multi}, recoverd using the multiscale approach. The geodesic errors are measured with respect to the ground-truth on the finest scale. At coarse scales the minimal expected geodesic error introduced by any matching is in the order of the sampling radius. As expected, the accuracy of the matching increases with each iteration.}
	\label{fig:cat_err_curve}
\end{figure}

In this set of experiments we demonstrate how the PMF together with the multiscale method described in \ref{sec:multi} can recover very accurate matchings on shapes being sampled at high resolution.
Figure \ref{fig:cats_multi} shows a dense bijective matching between two shapes sampled at $n=27894$ points each. At each of the six scales $n_i \in \{10^{3},2\times 10^3,4\times 10^3,8\times 10^3,1.6\times 10^4,n\}$ the constrained LAP \eqref{eq:multi_lap} was solved. Figure \ref{fig:cats_iters} shows the sequence of matchings over the scales. Figure \ref{fig:cat_err_curve} shows the improvement of correspondence accuracy at finer scales. By using the weighting functions we force points to stay close to their nearest neighbor in the coarser sampling and thus can guarantee to approximately keep the accuracy from the coarser scale. Solving the constrained LAP at the finest resolution took less then $9$ minutes. Calculating the kernel density matrices \eqref{eq:f_matrix} for all scales took less than $40$ minutes.

\begin{figure}	
	\begin{center}
		\includegraphics[width=0.95\columnwidth]{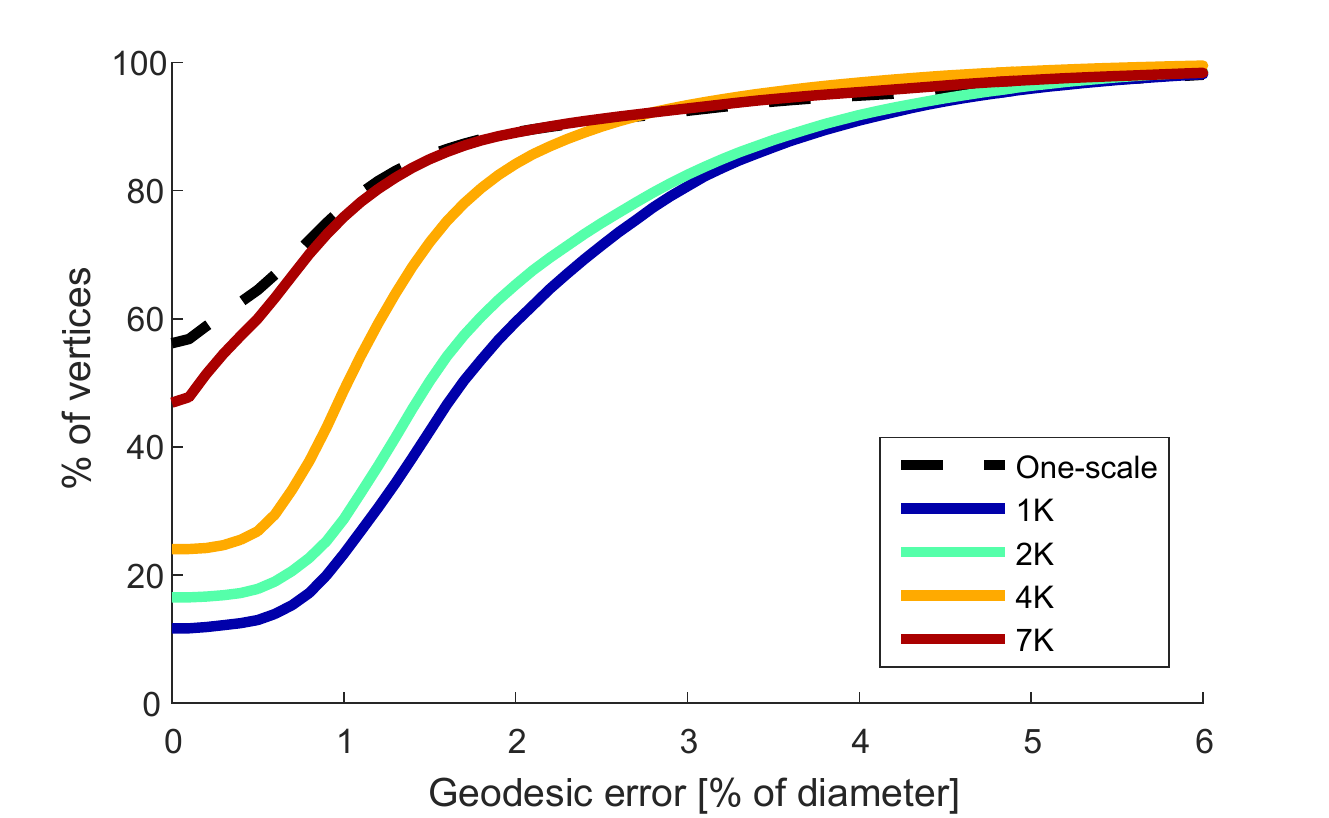} 
	\end{center}
	\caption{
		Error curves for a multiscale experiment on the FAUST dataset, showing result for intermediate scales.
		For comparison, the solution obtained by a single-scale PMF is shown in dashed black. 
	}
	\label{fig:faust_multi}
\end{figure}

Another test was performed on pairs of shapes from the FAUST dataset. As Figure~\ref{fig:faust_multi} shows, the correspondences obtained using a four-scale scheme are comparable in accuracy to the solution of a single-scale scheme. However, the runtime of the multi-scale approach is significantly lower.
Calculating the kernel density matrices \eqref{eq:f_matrix} for all scales took about $4$ minutes, while solving the LAPs at all scales took around $18$ seconds.

\begin{table}[]
\centering
\begin{tabular}{lcccc}
\hline\hline
 $n$& $1000$ & $1000$& $6890$& $6890$  \\
 $r$& $20$ & $50$ & $20$ & $50$\\
 \hline
 Nearest neighbors& 0.04 & 0.06 & 1.35 & 2.88 \\
 Bijective NN&2.79 & 2.30 & 463.66 & 253.03\\
 CPD&4.79 & 4.67 & 1745.06 & 2085.65 \\
 NN + PMF& 1.75 & 1.28 & 382.86 &244.10\\
 Bij. NN + PMF& 4.06 & 3.44 & 746.00 & 440.94 \\
  \hline \hline
\end{tabular}
\medskip
\caption{Average runtimes in seconds. We compare the runtimes of different correspondence recovery methods. Given the rank $k$ of a functional map approximating the correspondence between shapes sampled at $n$ points each, we report the time it takes to obtain a dense matching. See Figures \ref{fig:funmap_scape},\ref{fig:funmap_faust} and \ref{fig:funmap_qual} for evaluations of accuracy.}
\label{tab:run_times}
\end{table}

%% file: 5_conc.tex
\section{Discussion and conclusion}

We considered the problem of bijective correspondence recovery by means of filtering a given set of matches coming from any of the existing algorithms (including those not guaranteeing bijectivity, or producing sparse correspondences). Viewing correspondence computation as a kernel density estimation problem in the product space, we introduced the product manifold filter that leads to smooth correspondences, with the additional constraint of bijectivity embodied through an LAP.
We believe that statistical tools that have been heavily used in other domains of science and engineering might be very useful in shape analysis, and invite the community to further explore this direction. 
Of special interest is the possibility to lift the product space to higher dimensions encoding local similarity of points on the two shapes, for instance by using descriptors. The way the kernel density estimator is constructed does not restrict the samples per shape to be distinct. Together with the use of weighting factors this alllows to directly work with soft maps as inputs. 
Finally, we believe that denoising the correspondence manifold in the product space is a useful perspective applicable to different problems in computer vision where smooth correspondences are desired, such as optical flow.

\paragraph*{Acknowledgements}

We thank Zorah L\"ahner for useful discussions and Florian Bernard for the his mex implementation of the sparse auction algorithm. D.C. and M.V. are supported by the ERC CoG 3D Reloaded. A.B. is supported by the ERC StG RAPID. E.R. is supported by ERC StG No. 307047 (COMET). R.L. is supported by the European Google PhD fellowship in machine learning.

%% file: submission.bbl
\begin{thebibliography}{10}\itemsep=-1pt

\bibitem{aflalo2015optimality}
Y.~Aflalo, H.~Brezis, and R.~Kimmel.
\newblock On the optimality of shape and data representation in the spectral
  domain.
\newblock {\em SIAM Journal on Imaging Sciences}, 8(2):1141--1160, 2015.

\bibitem{aflalo2015convex}
Y.~Aflalo, A.~Bronstein, and R.~Kimmel.
\newblock On convex relaxation of graph isomorphism.
\newblock {\em PNAS}, 112(10):2942--2947, 2015.

\bibitem{SGMDS}
Y.~Aflalo, A.~Dubrovina, and R.~Kimmel.
\newblock Spectral generalized multidimensional scaling.
\newblock {\em IJCV}, 2016.

\bibitem{aflalo2013spectral}
Y.~Aflalo and R.~Kimmel.
\newblock Spectral multidimensional scaling.
\newblock {\em PNAS}, 110(45):18052--18057, 2013.

\bibitem{aigerman2016hyperbolic}
N.~Aigerman and Y.~Lipman.
\newblock Hyperbolic orbifold tutte embeddings.
\newblock {\em ACM Transactions on Graphics (TOG)}, 35(6):217, 2016.

\bibitem{aigerman2014lifted}
N.~Aigerman, R.~Poranne, and Y.~Lipman.
\newblock Lifted bijections for low distortion surface mappings.
\newblock {\em ACM Transactions on Graphics (TOG)}, 33(4):69, 2014.

\bibitem{SCAPE}
D.~Anguelov et~al.
\newblock {SCAPE}: Shape completion and animation of people.
\newblock {\em TOG}, 24(3):408--416, 2005.

\bibitem{WKS}
M.~Aubry, U.~Schlickewei, and D.~Cremers.
\newblock {T}he wave kernel signature: {A} quantum mechanical approach to shape
  analysis.
\newblock In {\em {P}roc. {ICCV}}, 2011.

\bibitem{bertsekas}
D.~P. Bertsekas.
\newblock {\em Network optimization: continuous and discrete models}.
\newblock Citeseer, 1998.

\bibitem{bes:mck:SURFACEMATCH}
P.~J. Besl and N.~D. McKay.
\newblock A method for registration of {3D} shapes.
\newblock {\em PAMI}, 14(2):239--256, 1992.

\bibitem{FAUST}
F.~Bogo, J.~Romero, M.~Loper, and M.~J. Black.
\newblock {FAUST}: {D}ataset and evaluation for {3D} mesh registration.
\newblock In {\em {P}roc. {CVPR}}, 2014.

\bibitem{Bronstein:2008}
A.~Bronstein, M.~Bronstein, and R.~Kimmel.
\newblock {\em Numerical Geometry of Non-Rigid Shapes}.
\newblock Springer, 2008.

\bibitem{bro:bro:kim:PNAS}
A.~M. Bronstein, M.~M. Bronstein, and R.~Kimmel.
\newblock Generalized multidimensional scaling: a framework for
  isometry-invariant partial surface matching.
\newblock {\em PNAS}, 103(5):1168--1172, 2006.

\bibitem{BronsteinK10}
M.~M. Bronstein and I.~Kokkinos.
\newblock Scale-invariant heat kernel signatures for non-rigid shape
  recognition.
\newblock In {\em Proc. CVPR}, 2010.

\bibitem{koltun}
Q.~Chen and V.~Koltun.
\newblock Robust nonrigid registration by convex optimization.
\newblock In {\em {P}roc. {ICCV}}, 2015.

\bibitem{ChenMedioni:91:ICP}
Y.~Chen and G.~Medioni.
\newblock Object modeling by registration of multiple range images.
\newblock In {\em Proc. Conf. Robotics and Automation}, 1991.

\bibitem{lafon:05:LOCAL}
R.~R. Coifman, S.~Lafon, A.~B. Lee, M.~Maggioni, B.~Nadler, F.~Warner, and
  S.~W. Zucker.
\newblock Geometric diffusions as a tool for harmonic analysis and structure
  definition of data: Diffusion maps.
\newblock {\em PNAS}, 102(21):7426--7431, 2005.

\bibitem{ela:kim:FLATTEN}
A.~Elad and R.~Kimmel.
\newblock Bending invariant representations for surfaces.
\newblock In {\em Proc. CVPR}, 2001.

\bibitem{lipman15}
I.~Kezurer, S.~Kovalsky, R.~Basri, and Y.~Lipman.
\newblock Tight relaxations of quadratic matching.
\newblock {\em Computer Graphics Forum}, 34(5), 2015.

\bibitem{KimLCF10}
V.~G. Kim, Y.~Lipman, X.~Chen, and T.~A. Funkhouser.
\newblock M{\"o}bius transformations for global intrinsic symmetry analysis.
\newblock {\em Computer Graphics Forum}, 29(5):1689--1700, 2010.

\bibitem{kim2011blended}
V.~G. Kim, Y.~Lipman, and T.~Funkhouser.
\newblock Blended intrinsic maps.
\newblock {\em TOG}, 30(4):79, 2011.

\bibitem{kovnatsky15}
A.~Kovnatsky, M.~M. Bronstein, X.~Bresson, and P.~Vandergheynst.
\newblock Functional correspondence by matrix completion.
\newblock In {\em Proc. CVPR}, 2015.

\bibitem{lahner16}
Z.~L\"ahner, E.~Rodol{\`a}, F.~R. Schmidt, M.~M. Bronstein, and D.~Cremers.
\newblock Efficient globally optimal 2d-to-3d deformable shape matching.
\newblock In {\em Proc. CVPR}, 2016.

\bibitem{leordeanu2005spectral}
M.~Leordeanu and M.~Hebert.
\newblock A spectral technique for correspondence problems using pairwise
  constraints.
\newblock In {\em Proc. ICCV}, 2005.

\bibitem{Lipman2011}
Y.~Lipman and I.~Daubechies.
\newblock {Conformal Wasserstein distances: Comparing surfaces in polynomial
  time}.
\newblock {\em Advances in Mathematics}, 227(3):1047 -- 1077, 2011.

\bibitem{litman2016spectrometer}
R.~Litman and A.~Bronstein.
\newblock Spectrometer: Amortized sublinear spectral approximation of distance
  on graphs.
\newblock In {\em 2016 International Conference on 3D Vision}, 2016.

\bibitem{Mateus08}
D.~Mateus, R.~P. Horaud, D.~Knossow, F.~Cuzzolin, and E.~Boyer.
\newblock Articulated shape matching using laplacian eigenfunctions and
  unsupervised point registration.
\newblock In {\em Proc. CVPR}, 2008.

\bibitem{Me11}
F.~M\'{e}moli.
\newblock {Gromov-Wasserstein Distances and the Metric Approach to Object
  Matching}.
\newblock {\em Foundations of Computational Mathematics}, pages 1--71, 2011.

\bibitem{Memoli:2005}
F.~M\'{e}moli and G.~Sapiro.
\newblock A theoretical and computational framework for isometry invariant
  recognition of point cloud data.
\newblock {\em Foundations of Computational Mathematics}, 5(3):313--347, 2005.

\bibitem{myronenko10}
A.~Myronenko and X.~Song.
\newblock Point set registration: Coherent point drift.
\newblock {\em TPAMI}, 32(12):2262--2275, 2010.

\bibitem{ovsjanikov2012functional}
M.~Ovsjanikov, M.~Ben-Chen, J.~Solomon, A.~Butscher, and L.~Guibas.
\newblock Functional maps: a flexible representation of maps between shapes.
\newblock {\em ACM Trans. on Graphics}, 31(4):30, 2012.

\bibitem{OvMe*10}
M.~Ovsjanikov, Q.~M\'{e}rigot, F.~M\'{e}moli, and L.~Guibas.
\newblock One point isometric matching with the heat kernel.
\newblock {\em Computer Graphics Forum}, 29(5):1555--1564, 2010.

\bibitem{pokrass2013sparse}
J.~Pokrass, A.~M. Bronstein, M.~M. Bronstein, P.~Sprechmann, and G.~Sapiro.
\newblock Sparse modeling of intrinsic correspondences.
\newblock In {\em Computer Graphics Forum}, volume~32, pages 459--468. Wiley
  Online Library, 2013.

\bibitem{rodola2012game}
E.~Rodol{\`a}, A.~M. Bronstein, A.~Albarelli, F.~Bergamasco, and A.~Torsello.
\newblock A game-theoretic approach to deformable shape matching.
\newblock In {\em Proc. CVPR}, 2012.

\bibitem{rodola16-partial}
E.~Rodol\`{a}, L.~Cosmo, M.~M. Bronstein, A.~Torsello, and D.~Cremers.
\newblock Partial functional correspondence.
\newblock {\em Computer Graphics Forum}, 2016.

\bibitem{rodola-vmv15}
E.~Rodol{\`a}, M.~Moeller, and D.~Cremers.
\newblock Point-wise map recovery and refinement from functional
  correspondence.
\newblock In {\em Proceedings Vision, Modeling and Visualization (VMV)},
  Aachen, Germany, 2015.

\bibitem{sahillioglu2012}
Y.~Sahillio\u{g}lu and Y.~Yemez.
\newblock Coarse-to-fine combinatorial matching for dense isometric shape
  correspondence.
\newblock {\em Computer Graphics Forum}, 30(5):1461--1470, 2011.

\bibitem{shtern2013matching}
A.~Shtern and R.~Kimmel.
\newblock Matching lbo eigenspace of non-rigid shapes via high order
  statistics.
\newblock {\em arXiv:1310.4459}, 2013.

\bibitem{solomon2012soft}
J.~Solomon, A.~Nguyen, A.~Butscher, M.~Ben-Chen, and L.~Guibas.
\newblock Soft maps between surfaces.
\newblock In {\em Computer Graphics Forum}, volume~31, pages 1617--1626, 2012.

\bibitem{tevs2011intrinsic}
A.~Tevs et~al.
\newblock Intrinsic shape matching by planned landmark sampling.
\newblock {\em Computer Graphics Forum}, 30(2):543--552, 2011.

\bibitem{torresani2008feature}
L.~Torresani, V.~Kolmogorov, and C.~Rother.
\newblock Feature correspondence via graph matching: Models and global
  optimization.
\newblock In {\em Proc. ECCV}, 2008.

\bibitem{umeyama1988eigendecomposition}
S.~Umeyama.
\newblock An eigendecomposition approach to weighted graph matching problems.
\newblock {\em PAMI}, 10(5):695--703, 1988.

\bibitem{kaick2010survey}
O.~van Kaick, H.~Zhang, G.~Hamarneh, and D.~Cohen-Or.
\newblock A survey on shape correspondence.
\newblock {\em Computer Graphics Forum}, 20:1--23, 2010.

\bibitem{Windheusericcv11}
T.~Windheuser, U.~Schlickewei, F.~R. Schmidt, and D.~Cremers.
\newblock Geometrically consistent elastic matching of 3d shapes: A linear
  programming solution.
\newblock In {\em IEEE International Conference on Computer Vision (ICCV)},
  2011.

\bibitem{ZaBo*09}
A.~Zaharescu, E.~Boyer, K.~Varanasi, and R.~Horaud.
\newblock Surface feature detection and description with applications to mesh
  matching.
\newblock In {\em Proc. CVPR}, 2009.

\end{thebibliography}
